\documentclass[10pt,twocolumn,letterpaper]{article}

\usepackage[utf8]{inputenc}
\usepackage[T1]{fontenc}
\usepackage{color}

\usepackage{algorithm}
\usepackage{algorithmicx}
\usepackage{algpseudocode}
\algloopdefx{Return}{\textbf{return} }
\makeatletter
\def\ALG@special@indent{%
    \ifdim\ALG@thistlm=0pt\relax
        \hskip-\leftmargin
    \else
        \hskip\ALG@thistlm
    \fi
}
\newcommand{\Statexx}{\item[]\noindent\ALG@special@indent}
\makeatother
\usepackage{listings}

\usepackage{bm}

\usepackage{preprint}
\usepackage{times}
\usepackage{epsfig}
\usepackage{graphicx}
\usepackage{amsmath}
\usepackage{amssymb}
\usepackage{url}
\usepackage{authblk}

\usepackage[pagebackref=true,breaklinks=true,letterpaper=true,colorlinks,bookmarks=false]{hyperref}

\begin{document}


\title{Deep Neural Network Learning with Second-Order Optimizers -- a Practical Study with a Stochastic Quasi-Gauss--Newton Method}

\renewcommand*{\Authfont}{\bfseries}
\author{Christopher Thiele\thanks{}\hspace{0.25in}Mauricio Araya-Polo\thanks{}\hspace{0.25in}Detlef Hohl\\
Shell International Exploration \& Production, Inc.\\
Houston, TX\\
}

\twocolumn[\begin{@twocolumnfalse}
\maketitle

\begin{abstract}
   Training in supervised deep learning is computationally demanding, and the convergence behavior is usually not fully understood.
   We introduce and study a second-order stochastic quasi-Gauss--Newton (SQGN) optimization method that combines ideas from stochastic quasi-Newton methods, Gauss--Newton methods, and variance reduction to address this problem.
   SQGN provides excellent accuracy without the need for experimenting with many hyper-parameter configurations, which is often computationally prohibitive given the number of combinations and the cost of each training process. 
   We discuss the implementation of SQGN with TensorFlow, and we compare its convergence and computational performance to selected first-order methods using the MNIST benchmark and a large-scale seismic tomography application from Earth science.
\end{abstract}
\vspace{0.5in}
\end{@twocolumnfalse}]
{
  \renewcommand{\thefootnote}%
    {\fnsymbol{footnote}}
  \footnotetext[1]{Christopher Thiele is a graduate student in the department of Computational and Applied Mathematics at Rice University. He contributed to the results presented in this paper as part of an internship at Shell International Exploration \& Production, Inc.}
  \footnotetext[2]{Mauricio Araya-Polo is now affiliated with Total E\&P R\&T S.A.}
}

  \section{Introduction}

  Weight optimization is, next to architecture design, the primary challenge in large-scale learning with deep neural networks, and it requires large computational resources.
  The problem combines four simultaneous challenges: large weight parameter dimension $m$, large data set dimension (number of labeled samples $n$ and size of each sample), non-convexity, and ill-posedness.
  While higher-order methods are in common use in the optimization community for subsets of these challenges, modified first-order stochastic gradient descent methods using backpropagation dominate in deep learning (see the extensive review by Bottou et al.~\cite{Bottou2018}).
  In this contribution, we revisit what makes first-order approaches so popular and how second-order methods can be applied in deep learning.
  We propose a stochastic quasi-Gauss-Newton (SQGN) method, a variant of stochastic quasi-Newton methods, as a particularly suitable approach, and we study its properties on a small (MNIST) and a large deep learning problem (seismic imaging tomography~\cite{Araya2018}).
  We show that SQGN can be applied to both problems with almost identical hyper-parameters.
 
  The paper is structured as follows:
  In section~\ref{sec:sgd_and_newton}, we revisit gradient descent and Newton's method and their application in the stochastic setting.
  We then review more advanced Newton-type methods in section~\ref{sec:literature}, and we discuss their suitability for the minimization of non-convex loss functions. Combining some of these ideas, we develop the SQGN method in section~\ref{sec:sqgn}, and we discuss its implementation with TensorFlow~\cite{Abadi2016} in section~\ref{sec:implementation}.
  In section~\ref{sec:numerical_experiments}, we apply SQGN to two deep neural network training problems and compare its convergence and computational performance to that of selected first-order methods.
  The paper concludes with a summary of our findings and an outlook on future work.

  \section{Gradient descent and Newton's method}\label{sec:sgd_and_newton}

  In this paper, we consider the problem of minimizing the empirical risk
  \begin{equation}\label{eq:empirical_risk}
    f=\frac 1n\sum_{i=1}^nf_i,
  \end{equation}
  where $f_1,\ldots,f_n:\mathbb R^m\rightarrow\mathbb R$.
  In the context of neural network training, each function $f_i$ takes the form
  \begin{equation}
    f_i(w)=L\left(h_i(w),y_i\right),
  \end{equation}
  where $h_i(w)$ is the output of the network when presented with the $i$th sample in the training data set using weights $w\in\mathbb R^m$, $y_i\in\mathbb R^r$ is the corresponding label, and $L$ is the loss function used to compare $h_i(w)$ and $y_i$.
  Throughout this paper, we follow the notation of Bottou et al.~\cite{Bottou2018} where possible.

  Gradient descent methods are perhaps the most common methods for the minimization of~(\ref{eq:empirical_risk}) in deep learning.
  In their simplest form, these methods use the iteration
  \begin{equation}\label{eq:gd_iteration}
    w_{k+1}=w_k-\alpha_k\nabla f(w_k),
  \end{equation}
  where the step sizes $\{\alpha_k\}$ are chosen such that the iterates $\{w_k\}$ converge to a local minimizer $w_\ast$ of $f$.
  Unfortunately, there is no natural scale for appropriate step sizes as the optimal choice depends on the scaling of $f$ itself.
  The sensitivity of the iteration to the scaling of $f$ can be avoided by using Newton's method, which determines the new iterate by finding a minimizer $d_k$ to the quadratic model
  \begin{equation}\label{eq:newton_model}
    \frac 12d^T\nabla^2f(w_k)d+\nabla f(w_k)^Td
  \end{equation}
  and setting $w_{k+1}=w_k+d_k$.
  For now, we assume that $\nabla^2f(w_k)$ is positive definite so that the minimizer $d_k$ exists and is the exact solution of the linear system
  \begin{equation}\label{eq:newton_update}
    \nabla^2f(w_k)d_k=-\nabla f(w_k).
  \end{equation}

  \subsection{Newton's method in the stochastic setting}\label{sec:newton_stochastic}

  Although Newton's method addresses the problem of step size selection, two separate difficulties arise when applying it in the context of neural network training:
  First, for a non-convex function $f$ the Hessian $\nabla^2f(w_k)$ can become indefinite, and Newton's method can converge to saddle points.
  Second, accurate gradients and Hessians of $f$ are often not available.
  For now, let us focus on the second issue.

  The computation of $f$, $\nabla f$, and $\nabla^2f$ requires evaluations of the individual functions $f_i$ and their gradients and Hessians.
  When $n$ and $m$ are large, these evaluations can be prohibitively expensive. 
  Instead, one may choose subsets or \emph{mini-batches} $\mathcal S_k,\mathcal T_k\subset\{1,\ldots,n\}$ and approximate $f$, $\nabla f$, and $\nabla^2f$ via
  \begin{equation}
    f_{\mathcal S_k} = \frac{1}{|\mathcal S_k|}\sum_{i\in\mathcal S_k}f_i,\ \nabla f_{\mathcal S_k} = \frac{1}{|\mathcal S_k|}\sum_{i\in\mathcal S_k}\nabla f_i,\label{eq:f_batch}
  \end{equation}
  and
  \begin{equation}
    \nabla^2f_{\mathcal T_k} = \frac{1}{|\mathcal T_k|}\sum_{i\in\mathcal T_k}\nabla^2f_i\label{eq:hessian_batch}.
  \end{equation}
  For simplicity, we restrict ourselves to the case where $\mathcal T_k\subset\mathcal S_k$.
  One optimizer iteration is then applied to $f_{\mathcal S_k}$ before a new mini-batch is selected.
  In the case of gradient descent, this strategy results in the well-established \emph{stochastic gradient descent} (SGD) method.
  Using mini-batches in Newton's method is more challenging, as it relies on the quadratic models~(\ref{eq:newton_model}):
  If these models are constructed with noisy gradients and Hessians, their minima may not converge to the minimum of the true loss function $f$.
  While the introduction of a decaying learning rate to the Newton updates can mitigate these difficulties, this somewhat conflicts with the idea of Newton's method.

  \section{Related work}\label{sec:literature}

  In this section, we review existing approaches to improve the convergence of Newton-type methods in cases where $f$ is stochastic or non-convex (see also Bottou et al.~\cite{Bottou2018}).

  \subsection{Regularization}

  Among the many techniques to address the issue of ill-posedness and non-convexity, shifting the eigenvalues of $\nabla^2f$ by some constant $\lambda>0$ is perhaps the simplest and most frequently used approach (see, e.g.,~\cite{Gower2016,Moritz2016,OLearyRoseberry2019,Kingma2014,Henriques2018}).
  If $\lambda$ is sufficiently large, the matrix $\nabla^2f+\lambda I$ is positive definite and can be used in place of the Hessian in the Newton update~(\ref{eq:newton_update}).
  Note that if $\nabla^2f$ is positive \emph{semi-definite}, any $\lambda>0$ ensures positive definiteness.

  \subsection{The Gauss--Newton method}

  For some loss functions, the semi-definite Gauss--Newton approximation can be used in place of the indefinite Hessian~\cite[section~6.3]{Bottou2018}.
  The classical Gauss--Newton method assumes that $f=\ell\circ h$, where
  \begin{equation}
    h:\mathbb R^m\rightarrow\mathbb R^{rn}, w\mapsto\begin{pmatrix}h_1(w)\\\vdots\\h_n(w)\end{pmatrix}
  \end{equation}
  are the concatenated network outputs and
  \begin{equation}
    \ell(h)=\frac 1n(h-y)^T(h-y)
  \end{equation}
  is the convex mean square error between $h$ and the labels~$y$.
  In this case, the Gauss--Newton approximation takes the form
  \begin{equation}
    H=\frac 2nJ_h^TJ_h\approx\nabla^2 f,
  \end{equation}
  where $J_h$ denotes the Jacobian of $h$ with respect to the weights $w$~\cite[section~2.4]{Kelley1999}.
  Note that $H$ is positive semi-definite by construction.

  The Gauss--Newton approximation can be extended to other functions $f$ that take the form $f=\ell\circ h$ for some convex function $\ell$.
  In these cases, the Gauss--Newton approximation is $H=J_h^T\nabla^2_h\ell J_h$, where $\nabla^2_h\ell$ is the (positive semi-definite) Hessian of $\ell$ with respect to the network output~$h$.
  In particular, this extension allows us to use the Gauss--Newton approximation for the cross-entropy loss function (see~\cite{Bottou2018,Schraudolph2002} and section~\ref{sec:mnist}).
  
  For Gauss--Newton-type optimization in neural network training, the generalized Gauss--Newton operator $H$ can be sampled (see sections~\ref{sec:newton_stochastic} and~\ref{sec:implementation}), or it can be approximated, e.g., using Kronecker-factored~\cite{Martens2015} or $\mathcal H$-matrix~\cite{Chen2019} approximations.

  \subsection{Line search and trust regions}

  Since the Newton update is computed based on a \emph{local} model of the loss function, a large step $d_k$ might lead to insufficient decrease in the value of $f$.
  Line search methods probe $f$ at different points along the line $w_k+\alpha d_k$ to ensure sufficient decrease~\cite{Kelley1999}.
  Trust region methods define a neighborhood $\mathcal N_k$ of $w_k$ in which the local model can be trusted.
  The update $d_k$ is then determined by \emph{constraining} the minimization of the model~(\ref{eq:newton_model}) to~$\mathcal N_k$, and the size of the trust region is adjusted based on a comparison of the actual decrease in the loss function and the decrease predicted by the model~\cite{Conn2000,Kelley1999}.
  The restriction $d_k\in\mathcal N_k$ ensures that that the update is well-defined even if the Hessian $\nabla^2f(w_k)$ is indefinite.

  Line search methods and trust regions have been applied in neural network training (see, e.g.,~\cite{OLearyRoseberry2019,Erway2019}), but both approaches can struggle in the stochastic case.
  In case of line search, sufficient decrease in $f_{\mathcal S_k}$ does not guarantee sufficient decrease in $f$.
  Similarly, replacing $f$ with its stochastic approximation $f_{\mathcal S_k}$ may affect the trustworthiness of a trust region model.
  When applying line search or trust regions in a stochastic setting, it may therefore be necessary to use large mini-batches~\cite{OLearyRoseberry2019} or mini-batches of increasing size~\cite{Erway2019}.

  \subsection{Inexact Newton and quasi-Newton methods}

  Computation of the Newton update $d_k$ by direct solution of the linear system~(\ref{eq:newton_update}) is often infeasible.
  Inexact Newton methods instead approximate $d_k$ using iterative solvers such as the conjugate gradient method~\cite{Kelley1999}.
  These methods rely on Hessian-vector products, which can be computed without forming the full Hessian (see section~\ref{sec:implementation}), and they have been successfully applied to machine learning problems~\cite{Martens2010,Martens2012,OLearyRoseberry2019}.
  Nonetheless, even the approximate solution of~(\ref{eq:newton_update}) can result in considerable computational cost for each Newton iteration.

  Quasi-Newton methods avoid this cost by using approximations $H_k\approx\nabla^2f(w_k)$ that are easily inverted.
  The classical LBFGS method constructs $H_k$ from weight updates $s_k=w_{k+1}-w_k$ and gradient differences
  \begin{equation}\label{eq:grad_diff}
    v_k=\nabla f(w_{k+1})-\nabla f(w_k)
  \end{equation}
  so that the inverse $H_k^{-1}$ can be applied using simple vector operations with the pairs $(s_{k-1},v_{k-1})$ (see, e.g.,~\cite{Kelley1999,Schraudolph2007}).
  LBFGS was first extended to stochastic optimization problems by Schraudolph et al.~\cite{Schraudolph2007}.
  More recently, Byrd et al.\ used sampled Hessian-vector products $v_k=\nabla^2f_{\mathcal T_k}(w_k)s_k$ instead of the gradient differences~(\ref{eq:grad_diff}) to improve the convergence of LBFGS for stochastic problems.
  This approach was further extended to include variance reduction~\cite{Moritz2016} (see below) and ideas from randomized linear algebra~\cite{OLearyRoseberry2019, Gower2016}.
  In this paper we present SQGN, a stochastic LBFGS-type method that uses Gauss--Newton matrix-vector products $v_k=J_{h_{\mathcal T_k}}^T\nabla^2_h\ell J_{h_{\mathcal T_k}}s_k$ instead of Hessian-vector products to ensure the positive (semi-)definiteness of the approximations $H_k$ to the Hessian for non-convex $f$.

  \subsection{Variance reduction}

  In order to make SGD more resilient to stochasticity, Johnson and Zhang~\cite{Johnson2013} proposed the use of stochastic variance-reduced gradients (SVRG) of the form
  \begin{equation}\label{eq:svrg_estimate}
     g_k=\nabla f_{\mathcal S_k}(w_k) - \nabla f_{\mathcal S_k}(\tilde w) + \nabla f(\tilde w)\approx\nabla f(w_k),
  \end{equation}
  where full gradients $\nabla f(\tilde w)$ are evaluated at certain intervals to improve upon the gradient approximation $\nabla f_{\mathcal S_k}(w_k)$ in each iteration.
  Combining SVRG with line search or trust region methods is difficult, as $g_k$ may not be a direction of descent for the sampled loss function $f_{\mathcal S_k}$, or it may not produce a good quadratic model for $f_{\mathcal S_k}$.
  Nonetheless, SVRG can be used with quasi-Newton methods which do not use line search or trust regions~\cite{Moritz2016}.

  \section{A stochastic quasi-Gauss--Newton method}\label{sec:sqgn}

  In this section, we present a stochastic quasi-Gauss--Newton (SQGN) method that combines ideas from Gauss--Newton methods, stochastic quasi-Newton methods, and SVRG (see section~\ref{sec:literature}).
  This combination makes SQGN similar to the method proposed by Moritz et al.~\cite{Moritz2016}, but it uses the Gauss--Newton approximation instead of the Hessian.
  Bottou et al.\ pointed out this approach as a way of applying stochastic quasi-Newton methods to non-convex problems~\cite[section~6.2.2]{Bottou2018}.
  Here, we show one such algorithm in detail before we discuss its implementation, convergence, and computational performance for a benchmark problem and an industrial application in later sections.

  Algorithm~\ref{alg:sqgn} describes the SQGN method step by step:
  Every $K$ iterations, the gradient $\mu$ of the full loss function is evaluated and stored along with the current weights~$\tilde w$ (lines~\ref{alg:sqgn:svrg_begin} to~\ref{alg:sqgn:svrg_end}).
  In each iteration, we select a mini-batch~$\mathcal S_k$ (line~\ref{alg:sqgn:batch_s}) and compute the sampled gradients $\nabla f_{\mathcal S_k}(w_k)$ and $\nabla f_{\mathcal S_k}(\tilde w)$ which in turn are used to compute a variance-reduced gradient $g_k$ (line~\ref{alg:sqgn:svrg}).
  We then determine the Newton step $d_k$ using the standard LBFGS procedure (line~\ref{alg:sqgn:lbfgs_direction}), which can be implemented efficiently using two loops over the history $\mathcal C$ of curvature pairs (see, e.g.,~\cite{Kelley1999,Schraudolph2007}).
  After the Newton step is scaled by a learning rate (lines~\ref{alg:sqgn:scale_begin} to~\ref{alg:sqgn:scale_end}), it is applied to the weights (line~\ref{alg:sqgn:weight_update}).
  Finally, we compute a new curvature pair $(s_k,v_k)$ using the Gauss--Newton operator (line~\ref{alg:sqgn:gn_application}) every $M$ iterations, and we upate the history of curvature pairs (lines~\ref{alg:sqgn:hist_update_begin} to~\ref{alg:sqgn:hist_update_end}).

  \begin{algorithm}[h]
    \caption{Stochastic quasi-Gauss--Newton method}
    \label{alg:sqgn}
    \begin{algorithmic}[1]
      \State $\mathcal C\leftarrow\emptyset$
      \For{$k=0,1,\ldots$}
        \If{$k\equiv 0$ (mod $K$)}\label{alg:sqgn:svrg_begin}
          \State $\mu\leftarrow\nabla f(w_k)$
          \State $\tilde w\leftarrow w_k$
        \EndIf\label{alg:sqgn:svrg_end}
        \State Select mini-batch $\mathcal S_k\subset\{1,\ldots,n\}$.\label{alg:sqgn:batch_s}
        \State $g_k=\nabla f_{\mathcal S_k}(w_k) - \nabla f_{\mathcal S_k}(\tilde w) + \mu$\label{alg:sqgn:svrg}
        \State Compute LBFGS direction $d_k$ from the curvature\label{alg:sqgn:lbfgs_direction}
        \Statexx history $\mathcal C$ (cf.\ lines~\ref{alg:sqgn:hist_update_begin}-\ref{alg:sqgn:hist_update_end} and~\cite[algorithm~3]{Schraudolph2007}).
        \If{$k=0$}\label{alg:sqgn:scale_begin}
          \State $\triangleright$ Start with a small first step (cf.~\cite{Schraudolph2007}).
          \State $s_k = 10^{-7}d_k$
        \Else
          \State $s_k = \alpha d_k$
        \EndIf\label{alg:sqgn:scale_end}
        \State $w_{k+1} = w_k + s_k$\label{alg:sqgn:weight_update}
        \If{$k\equiv 0$ (mod $M$)}
          \State Select mini-batch $\mathcal T_k\subset\mathcal S_k$.
          \State $v_k=J_{h_{\mathcal T_k}}^T\nabla^2_h\ell J_{h_{\mathcal T_k}}s_k$\label{alg:sqgn:gn_application}
          \State $\mathcal C\leftarrow\mathcal C\cup\{(s_k,v_k)\}$\label{alg:sqgn:hist_update_begin}
          \If{$|\mathcal C|>L$}
            \State Remove oldest curvature pair from $\mathcal C$.
          \EndIf\label{alg:sqgn:hist_update_end}
        \EndIf
      \EndFor
    \end{algorithmic}
  \end{algorithm}

  Note that Algorithm~\ref{alg:sqgn} uses \emph{two} approximations to the Hessian of $f$: the LBFGS approximation that is based on the history $\mathcal C$ of curvature pairs, and the Gauss--Newton approximation $J_{h_{\mathcal T_k}}^T\nabla^2_h\ell J_{h_{\mathcal T_k}}$ for the computation of the curvature pairs themselves.

  \section{Implementation}\label{sec:implementation}

  In the following, we describe a way of implementing the action of the Gauss--Newton operator using automatic differentiation (AD)~\cite{Bucker2006}.
  The implementation uses the identity
  \begin{equation}\label{eq:jacobian_action}
    \nabla(g^Tv)=J_g^Tv,
  \end{equation}
  where $J_g$ denotes the Jacobian of $g$, and $v$ is a constant vector.
  Equation~(\ref{eq:jacobian_action}) allows us to compute the action of the transposed Jacobian of $g$ using only the gradient of a scalar function, which is easily computed with libraries such as TensorFlow.
  If we set $g=\nabla f$ in Equation~(\ref{eq:jacobian_action}), we have $J_g^T=\nabla^2 f$, and hence we can compute Hessian-vector products as gradients of scalar functions as well (cf.~\cite[example~6.1]{Bottou2018}).

  In order to compute the action of the Gauss--Newton operator $J_h^T\nabla^2_h\ell J_h$, we use the procedure in Algorithm~\ref{alg:JTHJv}, which relies heavily on the identity~(\ref{eq:jacobian_action}) and on a technique to compute Jacobian-vector products with AD that was pointed out by Townsend~\cite{Townsend2017blog}.
  Algorithm~\ref{alg:JTHJv} allows the application of the Gauss--Newton operator using three passes of reverse mode AD or backpropagation.
  This distinguishes our approach from other techniques that use both forward mode and backward mode AD~\cite{Pearlmutter1994,Schraudolph2002,Henriques2018}.
  Our backpropagation-only approach is implemented more easily with libraries like TensorFlow, as it only requires gradients of scalar-valued functions.
  The computational performance of this approach depends on internal optimizations of the AD library.
  In particular, the library must realize that
  \begin{equation}
    \tilde\varphi(v)=\nabla_u(\varphi^Tv)=(J_h^T)^Tv=J_hv
  \end{equation}
  in Algorithm~\ref{alg:JTHJv} only has artificial dependencies to $u$ and $\varphi(u)=J_h^Tu$, as the final result $J_hv$ does not depend on either of them.\footnote{In a discussion (see \path{https://github.com/HIPS/autograd/pull/175}), Townsend and Johnson argue that in this case the cost of computing $J_hv=\tilde\varphi(v)$ via reverse mode AD is equivalent to that of computing $J_hv=\partial_vh$ as a directional derivative using forward mode AD (see~\cite{Pearlmutter1994,Schraudolph2002,Henriques2018}).}
  Finally, note that the second backpropagation pass, i.e., the evaluation of $\varphi(z)=\nabla_h^2\ell z$, does not require backpropagation through the neural network, as the Hessian is taken with respect to the network \emph{output} $h$.
  
  Our Python and TensorFlow-based SQGN implementation is available as open source software~\cite{Thiele2020sqgncode}.
  
  \begin{algorithm}[h]
    \caption{Computation of the action of the Gauss--Newton operator using backpropagation}
    \label{alg:JTHJv}
    \begin{algorithmic}[1]
      \State Let $\varphi:\mathbb R^{rn}\rightarrow\mathbb R^m,u\mapsto\nabla_w(h^Tu)$ and observe that $\varphi(u)=J_h^Tu$ according to Equation~(\ref{eq:jacobian_action}).
      \State Let $\psi:\mathbb R^{rn}\rightarrow\mathbb R^{rn},z\mapsto\nabla_h(\nabla\ell^Tz)$ and observe that $\psi(z)=\nabla^2\ell z$ according to Equation~(\ref{eq:jacobian_action}).
      \State Let $\tilde\varphi:\mathbb R^m\rightarrow\mathbb R^{rn},v\mapsto\nabla_u(\varphi^Tv)$ and observe that $\tilde\varphi(v)=(J_h^T)^Tv=J_hv$ (cf.~\cite{Townsend2017blog}).
      \State Compute $J_h^T\nabla^2\ell J_hv=\varphi(\psi(\tilde\varphi(v))$ using three passes of backpropagation.
    \end{algorithmic}
  \end{algorithm}

  \section{Numerical experiments}\label{sec:numerical_experiments}

  In this section, we evaluate the convergence behavior and computational performance of SQGN using two vastly different example problems: the well-understood small MNIST benchmark~\cite{LeCun2010} and a large-scale industrial seismic tomography application.
  For both problems, we compare the convergence and computational performance of SQGN against that of Adam~\cite{Kingma2014}.
  For reference, we also show results with a standard SGD method and with SVRG for selected configurations.

  All experiments were performed using a machine with two Intel Xeon Silver 4110 CPUs, 256~GB of memory, and a 16~GB PCIe NVIDIA V100 GPU (model no. 900-2G500-0000-000).
  The experiments were repeated five times to account for the stochasticity in the optimization process and spurious variations in system performance.

  \subsection{MNIST}\label{sec:mnist}

  The MNIST data set is a collection of images of handwritten digits~\cite{LeCun2010}.
  We use the default splitting of the data set into $60{,}000$ images for training and $10{,}000$ images for testing.
  In order to solve this image classification problem, we use a neural network with three convolutional layers and one dense layer.
  Each convolutional layer is followed by a ReLU activation function and max pooling, while the dense layer is followed by a softmax activation function and a cross-entropy loss function.
  The total number of trainable weights is $1{,}962$, and the weights are initialized using uniform Glorot initialization~\cite{Glorot2010}.

  \subsubsection{Baseline experiment}\label{sec:baseline}

  The first experiment compares the convergence of SGD, SVRG, Adam, and SQGN using the hyper-parameters below.
  We use these results as a reference for later experiments in which we investigate the effect of individual hyper-parameters in greater detail.

  For the baseline experiment, we train the neural network for $100$ epochs.
  In each epoch, the optimizer performs~$60$ iterations using mini-batches of size $|\mathcal S_k|=|\mathcal T_k|=1{,}000$ to sample the loss function, its gradient, and the Gauss--Newton operator.
  We still define an epoch as a sequence of $60$ optimizer iterations when additional gradients are computed for variance reduction.
  Hence, an epoch does not always correspond to a single pass over the training data, and we account for this by showing the average elapsed time per optimizer iteration along with the convergence results.
  For SGD, SVRG, and Adam, we use a learning rate of $\alpha=10^{-2}$, the optimal learning rate from the set $\{1, 10^{-1}, 10^{-2}, 10^{-3}\}$ for all three methods.
  For SQGN, we use a learning rate of $\alpha=10^{-1}$, which is close to, but more conservative than the natural rate of $\alpha=1$ for non-stochastic Newton-type methods.
  The interval for Gauss--Newton operator evaluations is $M=1$, and the history length is $L=20$ (see Algorithm~\ref{alg:sqgn}).
  The regularization parameter is $\lambda=10^{-1}$.
  SVRG-like variance reduction is used within SQGN, and full gradients are evaluated every $K=10$ iterations.

\begin{figure*}[t]
    \centering
    \includegraphics[width=0.4\textwidth]{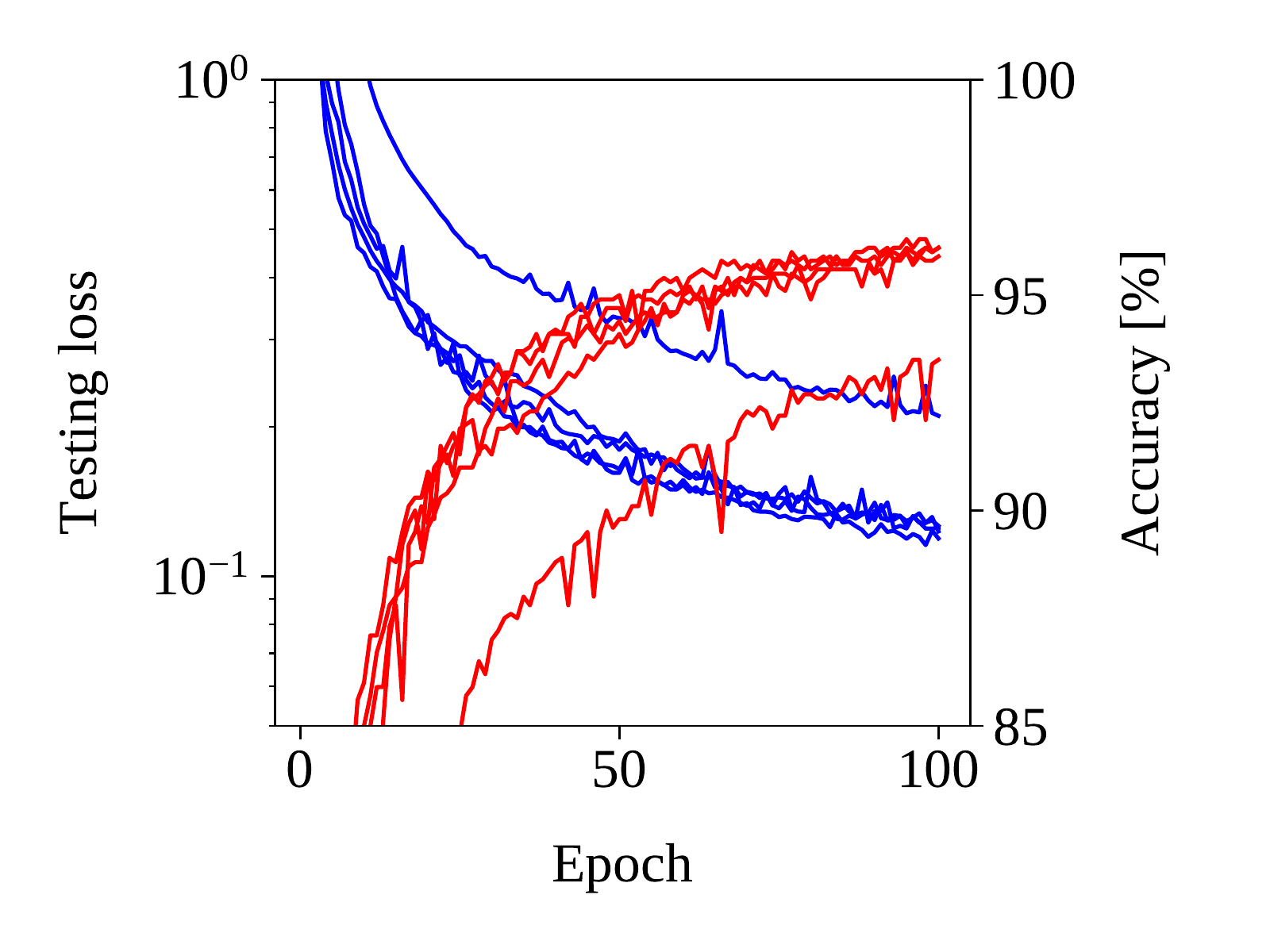}
    \includegraphics[width=0.4\textwidth]{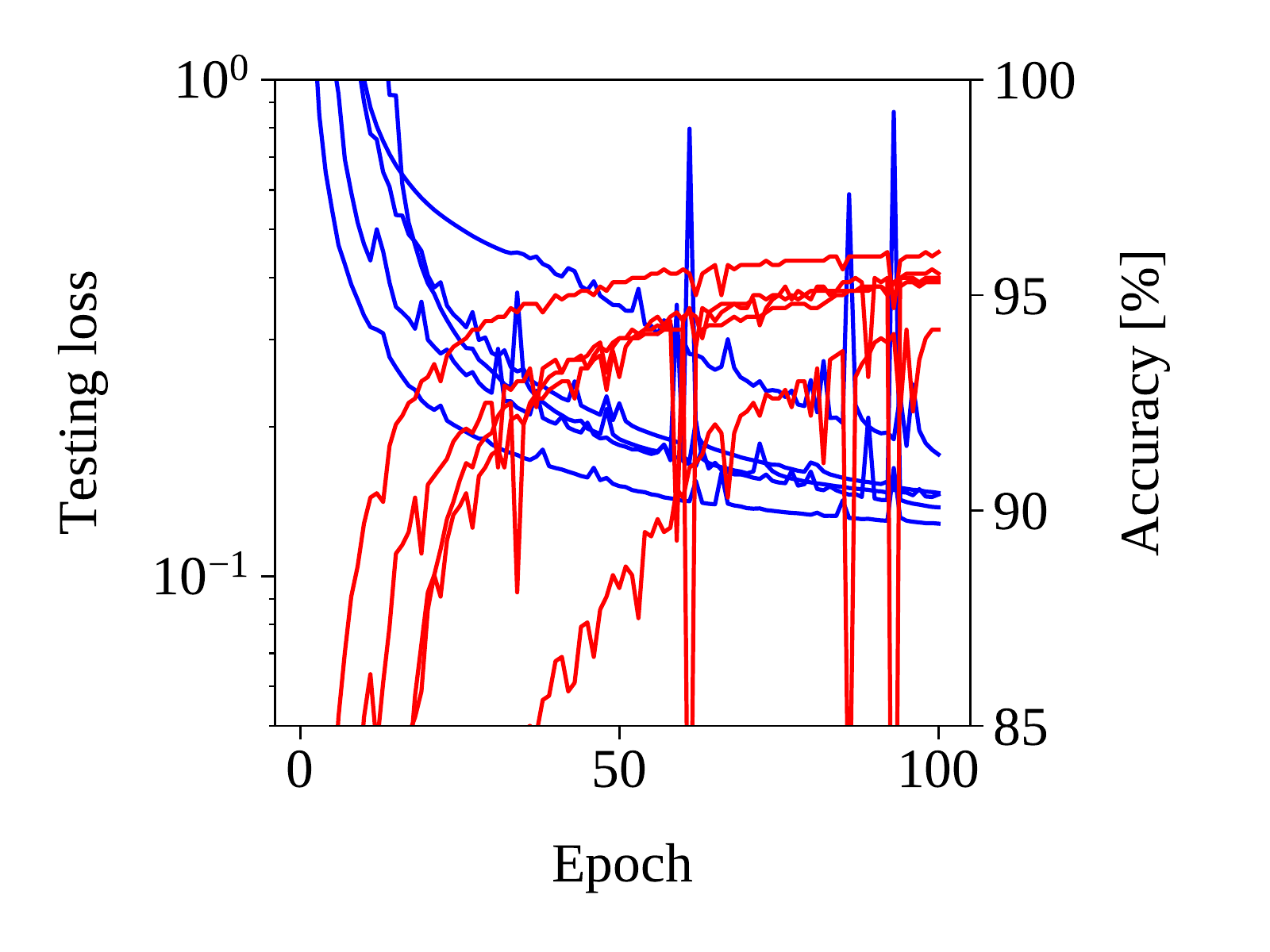}
    \includegraphics[width=0.4\textwidth]{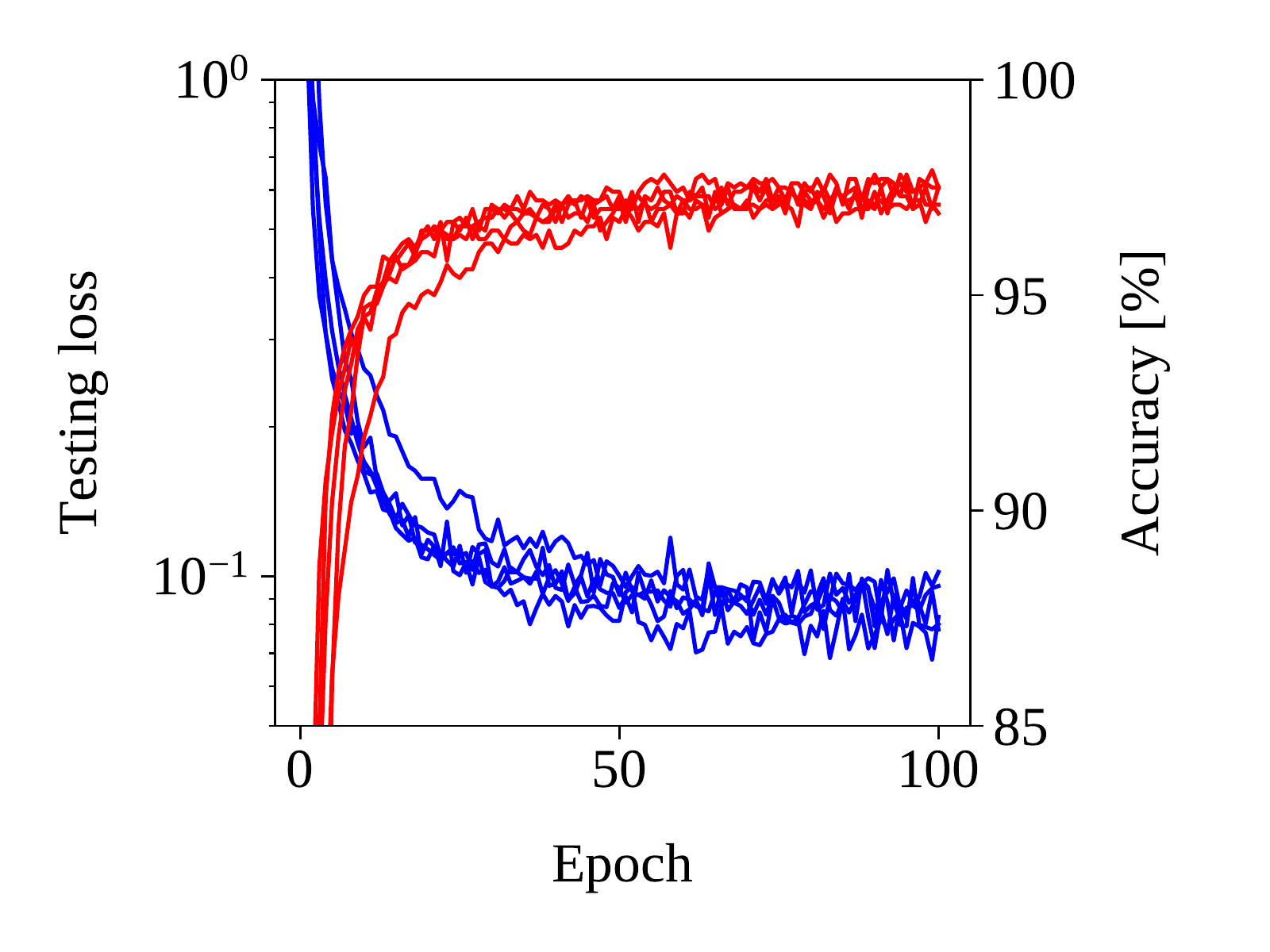}
    \includegraphics[width=0.4\textwidth]{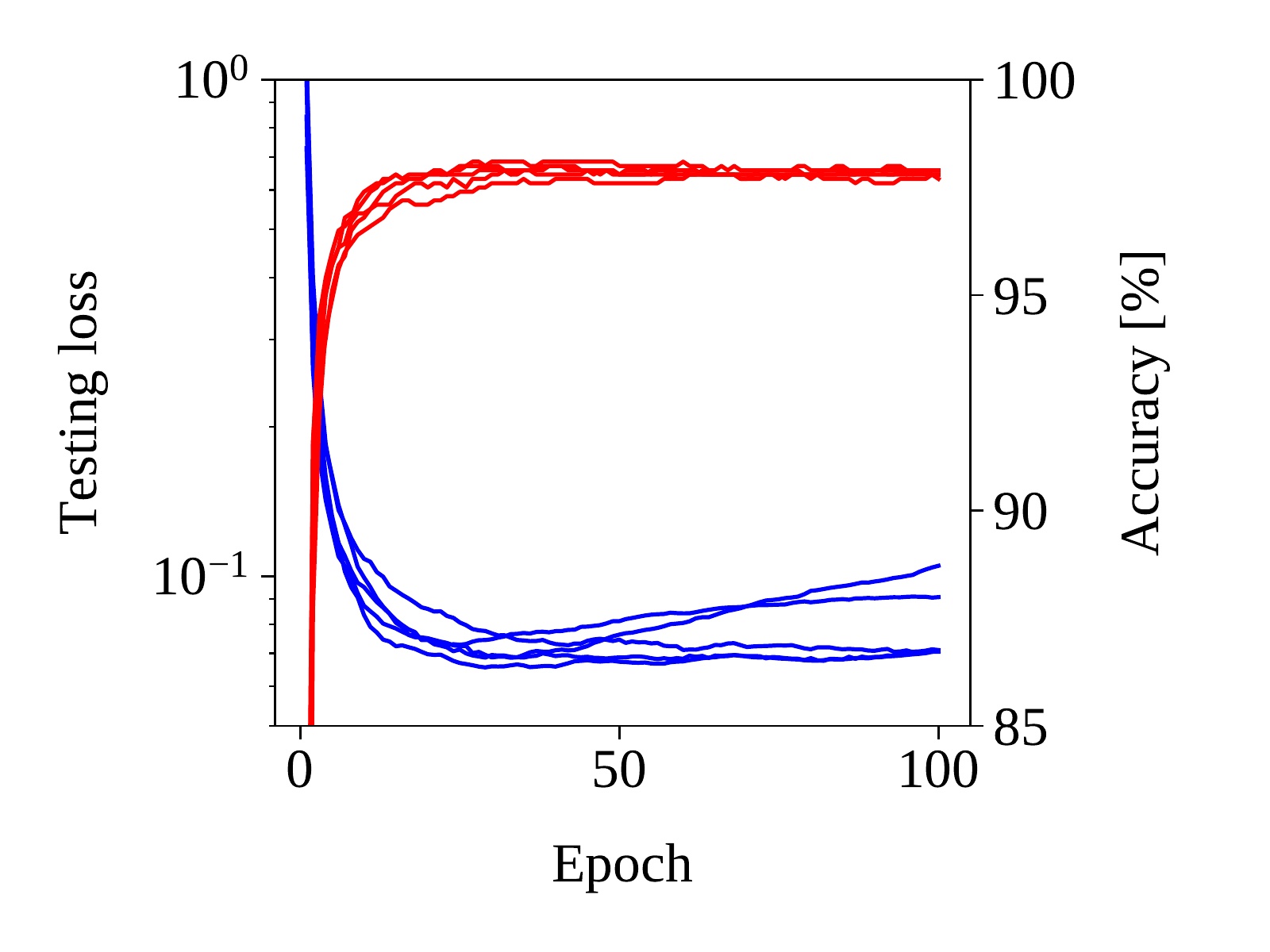}
    \caption{Convergence behavior for the baseline MNIST experiment with SGD (top left), SVRG (top right), Adam (bottom left), and SQGN (bottom right). Cross-entropy loss (blue) and classification accuracy (red) are plotted for five repetitions of the experiment.}
    \label{fig:baseline}
  \end{figure*}

  Table~\ref{tab:baseline} and Figure~\ref{fig:baseline} show the results of the baseline experiment.
  After $100$ epochs, SQGN achieves the lowest loss value and the highest accuracy, followed closely by Adam.
  In comparison, SGD and SVRG yield higher losses and lower accuracies after 100 epochs, although both methods will eventually reach accuracies of over $97\%$ if the number of epochs is increased.
  We acknowledge that the use of momentum or decaying learning rates can improve the convergence of SGD and SVRG and that such methods can be competitive with and in some cases preferable to Adam~\cite{Wilson2017}.
  Nonetheless, we focus on Adam as a reference for our evaluation of SQGN, and our main reason for showing results with SGD and SVRG is to illustrate that the rapid convergence of SQGN seen in Figure~\ref{fig:baseline} cannot simply be attributed to the occasional evaluation of full gradients for variance reduction.

  \begin{table}[h]
    \centering
    \caption{Final cross-entropy testing loss, classification accuracy, and average time per iteration for the MNIST baseline experiment after $100$ epochs with different optimizers. The reported losses, accuracies, and times are averaged over five repetitions of the experiment.}
    \begin{tabular}{@{\extracolsep{4pt}}lrrr}
      \hline
      \textbf{Method} & \multicolumn{1}{l}{\textbf{Loss}} & \multicolumn{1}{l}{\textbf{Accuracy}} & \multicolumn{1}{l}{\textbf{Avg. time/iter.}}\\
      \hline
      SGD  & $14.1\cdot 10^{-2}$ & $95.5\%$ &  $7$~ms\\
      SVRG & $14.7\cdot 10^{-2}$ & $95.3\%$ & $39$~ms\\
      Adam &  $8.8\cdot 10^{-2}$ & $97.3\%$ &  $7$~ms\\
      SQGN &  $8.2\cdot 10^{-2}$ & $97.8\%$ & $57$~ms\\
      \hline
    \end{tabular}
    \label{tab:baseline}
  \end{table}

  Table~\ref{tab:baseline} also shows the average elapsed time per iteration for the different optimizers.
  These times include the optimizer iteration itself and the transfer of training data to GPU memory.
  The average is taken over all $6{,}000$ iterations from $100$ epochs.
  For optimizers with variance reduction, the times include the cost of occasional full gradient evaluations.
  We see that the per-iteration cost of SGD and Adam, both gradient descent methods without variance reduction, is significantly lower than the per-iteration cost for SVRG and SQGN.
  Furthermore, SQGN is the more expensive of the two methods with variance reduction.
  This is expected, as the SQGN algorithm is significantly more complex than SVRG.

  The high per-iteration cost of SQGN is balanced to some degree by its convergence behavior:
  After $25$ epochs, SQGN achieves an average accuracy of $97.7\%$, which is already higher than the accuracies achieved with any of the three first-order methods after $100$ epochs (cf.\ Table~\ref{tab:baseline}).
  Once the accuracy becomes stationary after about $50$ SQGN iterations, the testing error increases again in some cases (see Figure~\ref{fig:baseline}).
  We interpret this as an overfitting of the neural network to the training data.

  While SQGN always achieves high accuracies of more than $97\%$ when it converges, the method sometimes gets stuck at accuracies of around $10\%$ or becomes unstable in early iterations.
  This behavior only occurred in about $3\%$ of our experiments, and it is easily detected after a few epochs.
  Hence, we simply restarted the training procedure when SQGN got stuck.
  Stability could perhaps be improved by taking more careful initial steps until a sufficient approximation to the Gauss--Newton operator has been constructed (see~\cite{Byrd2016}).
  It is worth mentioning that SGD also gets stuck at around $10\%$ accuracy in some cases.
  Thus, simply falling back to SGD in early SQGN iterations may not substantially improve stability.

  \subsubsection{Influence of the learning rate}\label{sec:learning_rate}

  The baseline experiment is repeated using diffent learning rates for both Adam and SQGN.
  Table~\ref{tab:learning_rates} shows the testing loss and accuracy after $100$ epochs.
  We see that the Adam optimizer fails if a large learning rate of $\alpha=10^{-1}$ is used.
  In comparison, SQGN is less sensitive to the choice of $\alpha$.
  When the learning rate is lowered to $\alpha=10^{-3}$, both methods converge more slowly, resulting in higher losses and lower accuracies after $100$ epochs.

  \begin{table}[h]
    \centering
    \caption{Convergence of SQGN and Adam for the MNIST problem with different learning rates $\alpha$. The reported cross-entropy testing losses and classification accuracies are averaged over five repetitions of the experiment.}
    \begin{tabular}{@{\extracolsep{4pt}}rrrrr}
      \hline
      & \multicolumn{2}{c}{\textbf{Adam}} & \multicolumn{2}{c}{\textbf{SQGN}}\\
      \cline{2-3}
      \cline{4-5}
      \multicolumn{1}{l}{$\bm{\alpha}$} & \multicolumn{1}{l}{\textbf{Loss}} & \multicolumn{1}{l}{\textbf{Acc.}} & \multicolumn{1}{l}{\textbf{Loss}} & \multicolumn{1}{l}{\textbf{Acc.}}\\
      \hline
      $10^{-1}$ & $230.3\cdot 10^{-2}$ & $11.1\%$ &  $8.2\cdot 10^{-2}$ & $97.8\%$\\
      $10^{-2}$ &   $8.8\cdot 10^{-2}$ & $97.3\%$ &  $7.4\cdot 10^{-2}$ & $97.9\%$\\
      $10^{-3}$ &   $9.9\cdot 10^{-2}$ & $96.9\%$ & $14.6\cdot 10^{-2}$ & $95.5\%$\\
      \hline
    \end{tabular}
    \label{tab:learning_rates}
  \end{table}

  \subsubsection{Influence of the mini-batch sizes and the Gauss--Newton operator evaluation interval}\label{sec:batch_sizes}

  The baseline experiment is repeated using different mini-batch sizes $|\mathcal S_k|$ and $|\mathcal T_k|$ and Gauss--Newton operator evaluation intervals $M$.
  Table~\ref{tab:batch_sizes} shows the testing loss and accuracy after $100$ epochs for SQGN and Adam.
  We see that SQGN achieves results similar to the baseline (first row) when the Gauss--Newton operator is evaluated less frequently ($M=10$, second row) or with smaller mini-batches ($|\mathcal T_k|=100$, third row).
  Even if the mini-batch size for gradient evaluations is reduced to $|\mathcal S_k|=100$ as well, SQGN's performance does not deteriorate (fourth row).
  We attribute this to the variance reduction within SQGN, as full gradients are evaluated every $K=10$ iterations regardless of the mini-batch sizes.

  In terms of computational performance, the results show that the mini-batch size $|\mathcal T_k|$ for Gauss--Newton operator evaluations has a limited impact, and that less frequent Gauss--Newton operator evaluations or smaller mini-batch sizes $|\mathcal S_k|$ for gradient computations offer greater improvements in performance.

  \begin{table*}[t]
    \centering
    \caption{Convergence and computational performance of SQGN and Adam for the MNIST problem with different mini-batch sizes. $|\mathcal S_k|$ and $|\mathcal T_k|$ are the mini-batch sizes for evaluations of the gradient and the Gauss--Newton operator respectively. $M$ is the interval at which the Gauss--Newton operator is evaluated (see Algorithm~\ref{alg:sqgn}). The reported cross-entropy testing losses and classification accuracies are averaged over five repetitions of the experiment.}
    \begin{tabular}{@{\extracolsep{4pt}}rrrrrrrrr}
      \hline
      &&& \multicolumn{3}{c}{\textbf{Adam}} & \multicolumn{3}{c}{\textbf{SQGN}}\\
      \cline{4-6}
      \cline{7-9}
      \multicolumn{1}{l}{$\bm{|\mathcal S_k|}$} & \multicolumn{1}{l}{$\bm{|\mathcal T_k|}$} & \multicolumn{1}{l}{$\bm{M}$} & \multicolumn{1}{l}{\textbf{Loss}} & \multicolumn{1}{l}{\textbf{Acc.}} & \multicolumn{1}{l}{\textbf{Avg. time/iter.}} & \multicolumn{1}{l}{\textbf{Loss}} & \multicolumn{1}{l}{\textbf{Acc.}} & \multicolumn{1}{l}{\textbf{Avg. time/iter.}}\\
      \hline
      $1{,}000$ &  $1{,}000$ &  $1$ &  $8.8\cdot 10^{-2}$ &      $97.3\%$ & $7$~ms        & $8.2\cdot 10^{-2}$ & $97.8\%$ & $57$~ms\\
      $1{,}000$ &  $1{,}000$ & $10$ &       \textquotedbl & \textquotedbl & \textquotedbl & $8.2\cdot 10^{-2}$ & $97.6\%$ & $49$~ms\\
      $1{,}000$ &    $100$ &  $1$ &       \textquotedbl & \textquotedbl & \textquotedbl & $7.7\cdot 10^{-2}$ & $97.7\%$ & $54$~ms\\
        $100$ &    $100$ &  $1$ & $17.8\cdot 10^{-2}$ &      $94.5\%$ & $3$~ms        & $7.9\cdot 10^{-2}$ & $97.5\%$ & $47$~ms\\
      \hline
    \end{tabular}
    \label{tab:batch_sizes}
  \end{table*}

  \subsubsection{Influence of variance reduction}

  When repeating the baseline experiment without SVRG-like variance reduction and with otherwise identical hyper-parameters, we observed that SQGN stagnates at higher loss values and lower accuracies of around $95\%$.
  There are also larger deviations in the achieved loss values and accuracies when variance reduction is disabled.
  We conclude that variance reduction is an integral part of the SQGN algorithm.

  Without variance reduction, the average time per SQGN iteration was $26$~ms compared to $57$~ms for the baseline experiment with variance reduction.
  Hence, variance reduction accounts for most of the computational cost of SQGN, and a careful choice of the interval at which full gradients are evaluated can help to improve SQGN's computational performance.

  \subsection{Seismic tomography}\label{sec:tomography}

\begin{figure}[h]
    \centering
    \includegraphics[width=\columnwidth]{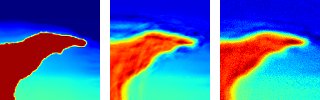}
    \caption{Example velocity model from the tomography problem, showing a salt body (red) in the subsurface. The figure shows the reference model (left) and the corresponding models predicted after training with Adam (center) and SQGN (right).}
    \label{fig:tomo3d_label_and_predictions}
  \end{figure}

  As our second numerical example, we use a seismic tomography problem with applications in hydrocarbon exploration and other inverse problems in the geosciences~\cite{Bergen19}.
  In seismic tomography, the task is to predict a subsurface model using data obtained from a seismic survey (see, e.g.~\cite{Nolet2008}).
  A machine learning-based approach to seismic tomography was described by Araya-Polo et al.~\cite{Araya2018}.
  We use a convolutional neural network (CNN) to predict \emph{velocity models}, i.e., earth models focused on a specific rock property, directly from 
  seismic data.
  Each velocity model is represented by an image of dimension $100\times 100$ that represents the speed of sound at different locations in the subsurface with a spacing of $15$ meters (see Figure~\ref{fig:tomo3d_label_and_predictions}).
  For $960$ such models, seismic surveys were simulated by propagating waves originating from $31$ sources and recording reflections at $256$ surface locations for $300$ time steps.
  The resulting seismic data of dimension $31\times 256\times 300$ per velocity model is then passed to the neural network, which attempts to recover the original models.

  The CNN has four three-dimensional convolutional layers with bias, each followed by a ReLU activation function and max pooling.
  The output of the last convolutional layer is passed into a dense layer, which is followed by a ReLU activation function and a mean square error loss function.
  The total number of trainable weights is $10{,}951{,}700$, and the weights are initialized using uniform Glorot initialization~\cite{Glorot2010}.
  Of the $960$ pairs of seismic data and velocity models, $800$ are used to train the neural network, and the remaining $160$ are used for testing.
  Mini-batches of size $|\mathcal S_k|=|\mathcal T_k|=100$ are used for training.

\begin{figure}[h]
    \centering
    \includegraphics[width=0.4\textwidth]{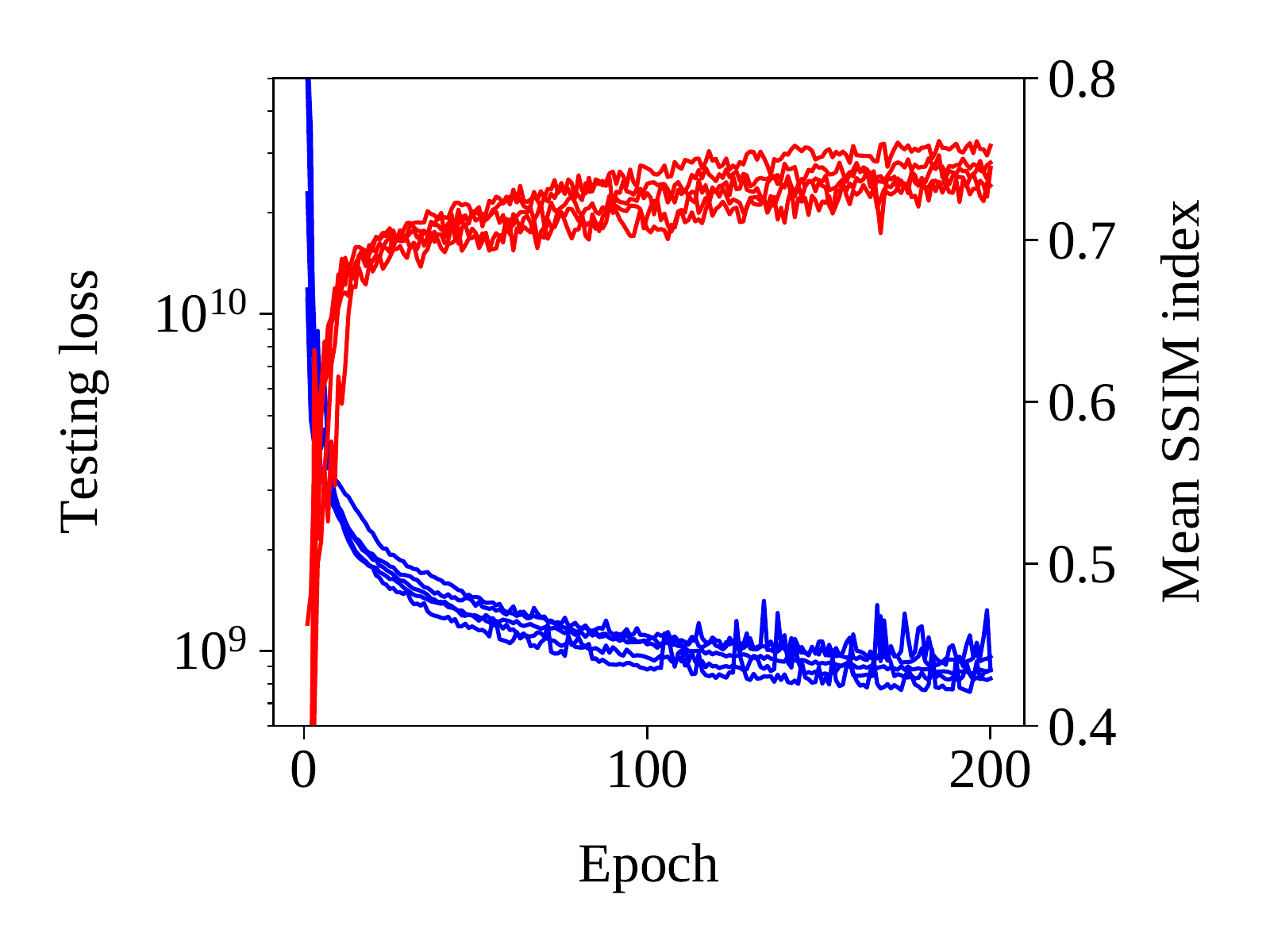}
    \includegraphics[width=0.4\textwidth]{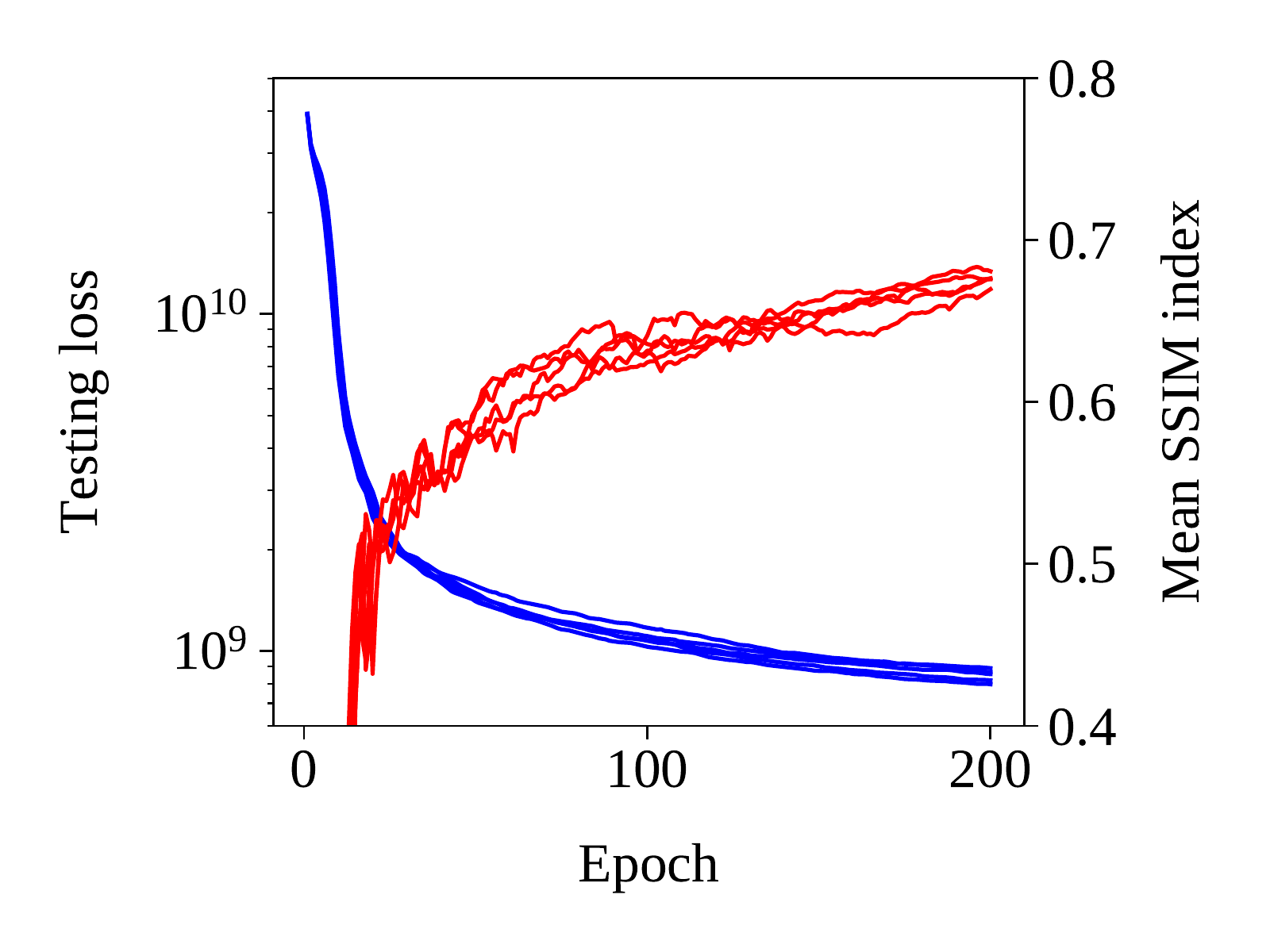}
    \caption{Convergence behavior of Adam ($\alpha=10^{-2}$, top) and SQGN (bottom) for the tomography problem. Loss (blue) and SSIM index (red) are plotted for five repetitions of the experiment.}
    \label{fig:tomo3d_convergence}
  \end{figure}

  Table~\ref{tab:tomo3d} and Figure~\ref{fig:tomo3d_convergence} show the convergence behavior of Adam and SQGN for the tomography problem.
  The results with SQGN were obtained using the same hyper-parameters $\alpha=10^{-1}$, $M=1$, $L=20$, and $\lambda=10^{-1}$ as for the MNIST problem in section~\ref{sec:baseline}.
  Hence, the results suggest that SQGN is insensitive to the choice of these parameters.
  Full gradients for variance reduction were evaluated once per epoch.

  On average, the Adam optimizer with $\alpha=10^{-2}$ and SQGN achieve similar losses, but there is less variance in the results obtained with SQGN.
  The variance in the results obtained with Adam can be reduced by using a smaller learning rate at the expense of slower convergence (see Table~\ref{tab:tomo3d}).

  In Table~\ref{tab:tomo3d} and Figure~\ref{fig:tomo3d_convergence}, we also report the mean structural similarity (SSIM) index~\cite{Wang2004}, where the mean is taken over all $160$ velocity models in the testing data set.
  The SSIM index is a measurement for the similarity of two images that is inspired by the human visual system.
  The maximum SSIM index is $1.0$, which implies that two images are identical.
  Our results show that the CNN achieves higher mean SSIM scores when it is trained with the Adam optimizer, even though the loss values are similar or greater than those obtained with SQGN.
  This is not a contradiction, since images with identical mean square error can have very different SSIM indices.
  In particular, the SSIM score is more sensitive to noisy images~\cite{Wang2004}, and as Figure~\ref{fig:tomo3d_label_and_predictions} shows, the predicted velocity models appear to be noisier when the CNN is trained using SQGN.
  Nonetheless, we do not consider the lower SSIM scores to be a weakness of SQGN, as the minimization problem is posed in terms of the mean square error, which SQGN reduces successfully.
  The target geological feature in Figure~\ref{fig:tomo3d_label_and_predictions} is the presence of salt (red structures).
  This salt formation is better recovered in the prediction by the SQGN-trained network.
  Furthermore, the prediction by the Adam-trained network has clearly visible artifacts, e.g., above the top left of the salt formation.

  For the tomography problem, the average per-iteration cost of SQGN is $4.5$~s, compared to $1.3$~s for Adam.
  While the difference in computational performance is still significant, it is less pronounced than for the MNIST problem in section~\ref{sec:mnist}.
  We attribute this improvement to the lower cost of variance reduction for the tomography problem:
  For the MNIST baseline experiment in section~\ref{sec:baseline}, the ratio $n/|\mathcal S_k|$ of the total number of samples in the training data set to the mini-batch size was greater than it is for the tomography problem ($n/|\mathcal S_k|=60{,}000/1{,}000=60$ for the MNIST problem vs.\ $n/|\mathcal S_k|=800/100=8$ here).
  Since we evaluated full gradients at similar intervals, the impact on the average per-iteration cost is reduced.

  \begin{table}[h]
    \centering
    \caption{Final testing loss, mean SSIM index, and average time per iteration for the tomography problem after $200$ epochs with different optimizers. The reported losses and SSIM indices are averaged over five repetitions of the experiment.}
    \begin{tabular}{@{\extracolsep{4pt}}lrrr}
      \hline
      \textbf{Method} & \multicolumn{1}{l}{\textbf{Loss}} & \multicolumn{1}{l}{\textbf{SSIM}} & \multicolumn{1}{l}{\textbf{Time/iter.}}\\
      \hline
      Adam ($\alpha=10^{-2}$) &  $8.7\cdot 10^8$ & $0.75$ & $1.3$~s\\
      Adam ($\alpha=10^{-3}$) & $10.0\cdot 10^8$ & $0.73$ & \textquotedbl\\
      SQGN                    &  $8.5\cdot 10^8$ & $0.68$ & $4.5$~s\\
      \hline
    \end{tabular}
    \label{tab:tomo3d}
  \end{table}

  \section{Conclusions and future work}

  We presented a stochastic quasi-Gauss-Newton (SQGN) method and applied to the problem of deep neural network training.
  The method incorporates ideas from existing stochastic quasi-Newton methods, and it uses the Gauss-Newton approximation to the Hessian to make it more suitable for non-convex optimization problems.
  We discussed the implementation of the method with state-of-the art software frameworks such as TensorFlow.
  We evaluated the convergence and computational performance of SQGN using two example problems of different complexity:
  handwritten digit classification (MNIST) and an industrial seismic imaging application.
  Our results show that SQGN converges in both examples using almost identical hyper-parameters, achieving results comparable to or better than those obtained with Adam.
  SQGN is robust with respect to the choice of learning rate, mini-batch sizes, and other hyper-parameters.
  We found that variance reduction is essential to ensure the convergence of SQGN to low loss function values.
  Finally, we identified optimizations that could make SQGN more competitive with popular first-order methods in terms of computational performance.

  Future work may include an evaluation of SQGN with supersampled or randomly sampled Gauss-Newton operators, motivated by successful applications of such techniques to stochastic quasi-Newton methods~\cite{Byrd2016,Martens2010,Gower2016}.
  SQGN can also be implemented with multi-GPU support in order to evaluate its convergence and performance for even larger problems.

  \section{Acknowledgments}

  The authors thank Shell International Exploration \& Production, Inc. for permitting the publication of the material presented in this paper.

{\small
\bibliographystyle{ieee_fullname}
\bibliography{bibliography}
}

\end{document}